\documentclass[onecolumn]{article}%
\usepackage{amsmath}
\usepackage{amsfonts}
\usepackage{amssymb}
\usepackage{graphicx}%
\providecommand{\U}[1]{\protect\rule{.1in}{.1in}}
\begin{document}

\title{An experimental mechatronic design and control of a 5 DOF Robotic arm for
identification and sorting of different sized objects}
\author{Christos Tolis and George F. Fragulis\\Laboratory of Robotics and Applied Control Systems \\Dept. Of Electrical Engineering\\Western Macedonia Univ. of Applied Sciences ,Kozani, Hellas }
\maketitle

\begin{abstract}
The purpose of this paper is to present the construction and programming of a
five degrees of freedom robotic arm which interacts with an infrared sensor
for the identification and sorting of different sized objects. The main axis
of the construction design will be up to the three main branches of science
that make up the Mechatronics: Mechanical Engineering, Electronic-Electrical Engineering and Computer Engineering. The methods that have been used for the construction are presented as well as the methods for
the programming of the arm in cooperation with the sensor. The aim is to
present the manual and automatic control of the arm for the recognition and
the installation of the objects through a simple (in operation) and low in
cost sensor like the one that was used by this paper. Furthermore, this paper
presents the significance of this robotic arm design and its further
applications in contemporary industrial forms of production.

\end{abstract}


\section{Introduction}

The aim of this paper is to  present the control and driving mechanism of robotic arm, which aims to grab different sized objects which can find applications within \ the industry or other working environments. The robotic arm must be highly functional, light weight and provide ease of attachment and control. 
The interdisciplinarity that characterizes Mechatronics between the notions of
Mechanical Engineering, Electronic-Electrical Engineering and Computer Science
is used for the selection of materials and devices to construct the arm. Furthermore it helped us to encounter any kinetic, power, torque, compatibility or other problems that would have resulted from the completion of the project . 
The major problem these robotic arms face is their cost (\cite{Lee2017}, \cite{Barakat2016}, \cite{Khanna2016}). The main factors for their expensiveness are use of advanced actuators, too complex design and manufacturing techniques and finnaly specialized sensors  for user input and control. To address the challenge we adopted a design that uses infrared(IR) sensors providing the virtual vision and low cost commercially available actuators.

\section{\bigskip Robot Arm and Infrared Sensor Description}

The Robotic Arm is a modular arm consisting of five rotary joints plus the end
effector which is a grip. The five rotating joints consist of: 1 joint for
base rotation, 1 for shoulder rotation, 1 for elbow rotation, 1 for wrist
rotation and 1 for grip rotation. The mechanical parts of the project were
selected by Lynxmotion one by one to meet our needs and are of the AL5 type.
The six servo motors by Hitec are chosen based on their torque for proper
operation. The Infrared Sensor is by Sharp and is a distance sensor. The Sharp
2Y0A21 F46 composed of an integrated combination of PSD (position sensitive
detector), IRED (infrared emitting diode) and signal processing circuit. The
device outputs the voltage corresponding to the detection distance.

Every rotational joint of the arm is being controlled by servo motors. These
motors are connected to a microcontroller (BotBoarduino) which is controlled
by a computer. The sensor is placed in front of the arm, above gripper, so can
`'read\textquotedblright\ the distance between the gripper and the reflected surface.

\section{Mechanical Engineering issues}

We studied the torque of the servo motors which have been chosen, to avoid any
kinetic problems. Furthermore we analyze the Degrees Of Freedom 
and the Work Space of the arm. To control the arm, the Forward
Kinematics  and Inverse Kinematics  has also been developed. After
we measured all the materials of the arm we made the CAD through the
SolidWorks software

\subsection{Torque Calculation}

Torque ($T$) is defined (\cite{serway},\cite{benson}) as a turning or twisting force and is calculated using the relation:%

\[
T=F\ast L
\]

The force ($F$) acts at a length ($L$) from a pivot point. In a vertical plane
the force that causing an object to fall is the acceleration due to gravity
($g$) multiplied by its mass ($m$):%

\[
F=g\ast m
\]

The above relation is the object's weight ($W$):%

\[
W=m\ast g
\]

\begin{figure}[!h]
	\centering
	\includegraphics[scale=0.5]{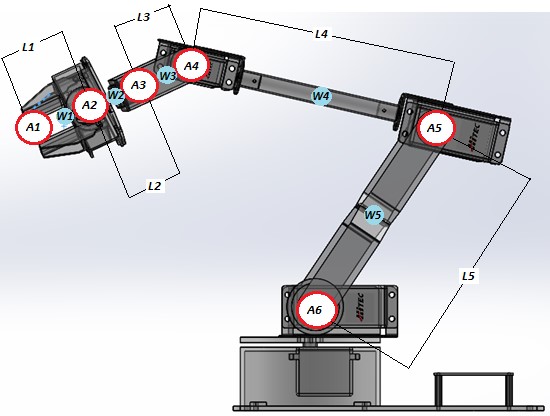}
	\caption{}
	\label{fig:3.1.1}
\end{figure}

In Figure \ref{fig:3.1.1}  we can see the lengths ($L$) of the links as well as the
weights ($W$) of the links considered that the center of mass is located at
roughly the center of its length. The $a_{1}$ in the image is the
`'load\textquotedblright\ being held by the arm, the $a_{2},a_{3},a_{4},a_{5}$
and $A6$ are the actuators (servos). To calculate the required torque ($T6$)
of the $A6$ motor ($HS-805BB$ servo) we use the relation:%

\begin{align}
T_{6}  &  =(L_{1}+L_{2}+L_{3}+L_{4}+L_{5})\ast A_{1}+(0.5\ast L_{1}%
+L_{2}+L_{3}+L_{4}+L_{5})\ast W_{1}\nonumber\\
&  +(L_{2}+L_{3}+L_{4}+L_{5})\ast A_{2}+(0.5\ast L_{2}+L_{3}+L_{4}+L_{5})\ast
W_{2}\label{3.1.1_0}\\
&  +(L_{3}+L_{4}+L_{5})\ast A_{3}+(0.5\ast L_{3}+L_{4}+L_{5})\ast
W_{3}\nonumber\\
&  +(L_{4}+L_{5})\ast A_{4}+(0.5\ast L_{4}+L_{5})\ast W_{4}+(L_{5})\ast
A_{5}+(0.5\ast L_{5})\ast W_{5}\nonumber
\end{align}

To calculate the required torque ($T5$) of the $a_{5}$ motor ($HS-755HB$ servo):%

\begin{align}
T_{5}  &  =(L_{1}+L_{2}+L_{3}+L_{4})\ast A_{1}+(0.5\ast L_{1}+L_{2}%
+L_{3}+L_{4})\ast W_{1}\nonumber\\
&  +(L_{2}+L_{3}+L_{4})\ast A_{2}+(0.5\ast L_{2}+L_{3}+L_{4})\ast
W_{2}\label{3.1.2}\\
&  +(L_{3}+L_{4})\ast A_{3}+(0.5\ast L_{3}+L_{4})\ast W_{3}+(L_{4}+L_{5})\ast
A_{4}+(0.5\ast L_{4}+L_{5})\ast W_{4}\nonumber
\end{align}

The torque ($T4$) of the $a_{4}$ ($HS-645MG$ servo) is calculated:%

\begin{align}
T_{4}  &  =(L_{1}+L_{2}+L_{3})\ast A_{1}+(0.5\ast L_{1}+L_{2}+L_{3})\ast
W_{1}\label{3.1.3}\\
&  +(L_{2}+L_{3})\ast A_{2}+(0.5\ast L_{2}+L_{3})\ast W_{2}+(L_{3})\ast
A_{3}+(0.5\ast L_{3})\ast W_{3}\nonumber
\end{align}

\bigskip In a same manner the torque ($T3$) of the $a_{3}$ ($HS-225MG$ servo):%

\begin{equation}
T_{3}=(L_{1}+L_{2})\ast A_{1}+(0.5\ast L_{1}+L_{2})\ast W_{1}+(L_{2})\ast
A_{2}+(0.5\ast L_{2})\ast W_{2} \label{3.1.4}%
\end{equation}

and finally the torque ($T2$) of the $a_{2}$ ($HS-422$ servo):%

\begin{equation}
T_{2}=(L_{1})\ast A_{1}+(0.5\ast L_{1})\ast W_{1} \label{3.1.5}%
\end{equation}

where: \ $A_{2}=45.5g(HS-422)$, $A_{3}=31g(HS-225MG)$, \ $A_{4}%
=55.2g(HS-645MG)+7g(ASB-24)=62.2g$, $A_{5}=110g(HS-755HB)+13g(ASB-201)=123g$,
$A_{6}=197g(HS-805BB)+18g(ASB-204)=215g$ .

The weights are \ $W_{1}=15.7g$ (Grip), $W_{2}=10g$ (Sensor Bracket),
$W_{3}=9g$(Wrist Bracket), $W_{4}=10g$ (AT-04) +$6g$(ASB-06)+ $8g$ (HUB-08)
=$24g$, $W_{5}=16g$ (ASB-205)+$15g$ ( ASB-203) = $31g$.

The lengths are : $L_{1}=2.8$ cm, $L_{2}=2.8$ cm, $L_{3}=2.85$cm,
$L_{4}=18.73$ cm, $L_{5}=14.6$ cm. \ If we replace the above, to the relations
(\ref{3.1.1_0}) - (\ref{3.1.5}) and the weight of $A_{1}$ (load) is zero, the
torques of the motors are:

\begin{itemize}
\item $T_{1}=0.021$ kg/cm

\item $T_{2}=0.207$ kg/cm

\item $T_{3}=0.511$ kg/cm

\item $T_{4}=5.122$ kg/m

\item $T_{5}=12.25$kg/cm
\end{itemize}

The nominal torques of the servo motors as given by the manufacturer are:

\begin{itemize}
\item $HS-422(T_{1})=4.1$ kg/cm

\item $HS-225MG(T_{2})=4.8$ kg/cm

\item $HS-645MG(T_{3})=9.6$ kg/cm

\item $HS-755HB(T_{4})=13.2$ kg/cm

\item $HS-805BB(T_{5})=24.7$ kg/cm
\end{itemize}

\bigskip From the above, we can say that the arm is capable to lift its own
weight because the nominal torques of the servos overlap the calculated
torques with zero load ($A_{1}=0$). \ Now If the load $A_{1}$ is set to be
$100g$ the torques are:

\begin{itemize}
\item $T_{1}=0.3$ kg/cm

\item $T_{2}=0.767$ kg/cm

\item $T_{3}=1.356$ kg/cm

\item $T_{4}=7.84$kg/cm

\item $T_{5}=16.43$ kg/cm
\end{itemize}

We notice, if the load is $A_{1}=100g$, servos can cope. If we increase the
weight at $A_{1}=300g$, then the calculated torques are:

\begin{itemize}
\item $T_{1}=0.86$ kg/cm

\item $T_{2}=1.887$ kg/cm

\item $T_{3}=3.04$ kg/cm

\item $T_{4}=13.27$ kg/cm

\item $T_{5}=24.79$ kg/cm
\end{itemize}

Hence the maximum weight the arm can lift is approximately \symbol{126}300g
because the calculated torques reach the nominal torques of the servos.

\bigskip

\subsection{DOF (Degrees Of Freedom)}

The arm has six actuators, one of which is to open and close the gripper, thus
not being considered as a degree of freedom. The five rotational actuators
have a degree of freedom each, with a result that the hole system has a total
of five degrees of freedom. If we see the DOF of the arm through a
mathematical point of view, the equation describing it is the
Gruebler-Kutzbach equation and is expressed by the relation:%
\[
M=3\ast(n-1)-2\ast J_{1}-J_{2}%
\]

where $M$ is the DOF of the system, $n$ is the number of links including the
base frame, $J_{1}$ is the number of joints that have one DOF, $J_{2}$ is the
number of joints that have more than one DOF

\begin{figure}[!h]
	\centering
	\includegraphics[scale=0.5]{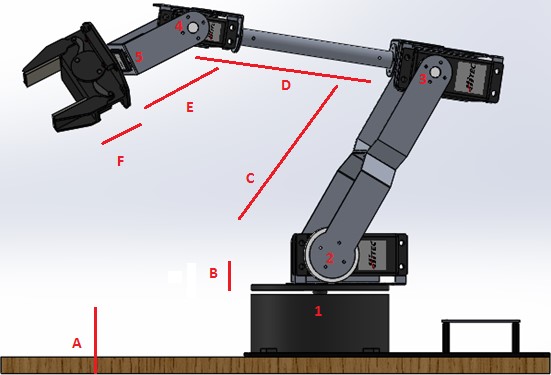}
	\caption{}
	\label{image312}
\end{figure}

In figure \ref{image312}  we can see the links (A, B, C, D, E and F) and the
joints (1, 2, 3, 4 and 5) of the arm. The number of the $J_{2}$ is zero
because there is no joints of two DOF in the system.

Therefore:%
\begin{align}
M  &  =3\ast(n-1)-2\ast J_{1}-J_{2}=3\ast5-10\Rightarrow\label{3.2.3}\\
M  &  =5\text{ DOF}\nonumber
\end{align}

\subsection{Work Space}

\begin{figure}[!h]
	\centering
	\includegraphics[scale=0.5]{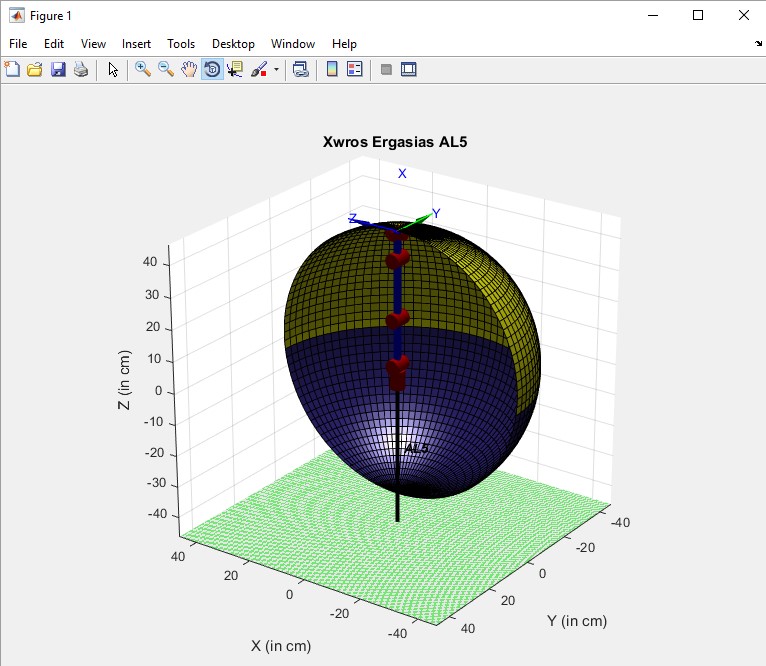}
	\caption{}
	\label{image313}
\end{figure}

The work space of the arm is the space where the end effector can act. We
compiled a code using MatLab for the 3D representation of the arm's work
space. In Figure \ref{image313} we can see the 3D presentation of the work space
of the arm using the Robotics toolbox\texttrademark (see \cite{Corke1996}, \cite{Corke2011}. The top hemisphere (colored: yellow) is the actual work space of
the arm and the bottom hemisphere (colored: blue) is a possible work space of
the arm under certain circumstances. The diameter of the work area of the arm
is approximately 40 cm .

\bigskip

\subsection{Forward Kinematic}

\begin{figure}[!h]
	\centering
	\includegraphics[scale=0.5]{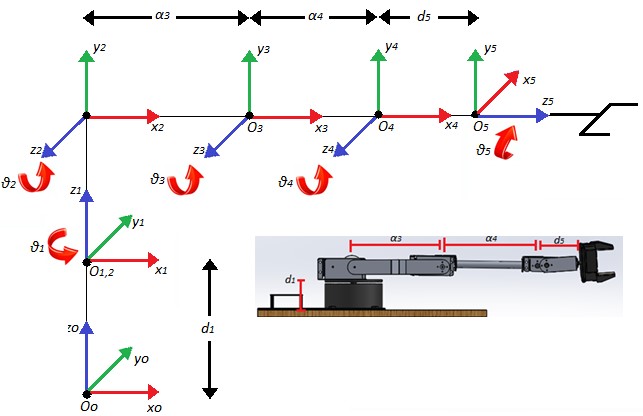}
	\caption{}
	\label{image314}
\end{figure}

Forward Kinematics (\cite{spong}, \cite{Graig2017})refers to the use of kinematics equations of a robot to
calculate the position of the end-effector from specified values for the joint
parameters. The Denavit-Hartenberg parameters is the most common method being
used to determine the Forward Kinematics analyses. Using this method we define
the coordinate frames of the arm (Fig. \ref{image314}), depending of the joints of
mechanisms and then the D-H parameters table (Table 1) has been calculated

Coordinate Frames has been defined with respect to D-H methodology where:

a$_{i}$ is the length of the common perpendicular between points $O_{1,2}$ and
$O_{4}$, $\ \ \alpha_{i}$ is the angle between axes $z_{i}$ and $z_{i-1},$
$d_{i}$ is the displacement distance of points $O_{0}-O_{1,2}$ and
$O_{4}-O_{5},$ $\theta_{i}$ is the angle between axis $x_{i}$ and $x_{i-1}.$%

\begin{equation}
\left[
\begin{array}
[c]{ccccc}%
\text{\textbf{Link i}} & \mathbf{a}_{i} & \mathbf{\alpha}_{i} & \mathbf{d}_{i}
& \mathbf{\theta}_{i}\\
1 & 0 & 0 & d_{1} & \theta_{1}\\
2 & 0 & 90^{o} & 0 & \theta_{2}\\
3 & \text{a}_{3} & 0 & 0 & \theta_{3}\\
4 & \text{a}_{4} & 0 & 0 & \theta_{4}(90^{o})\\
5 & 0 & -90^{o} & d_{5} & \theta_{5}%
\end{array}
\right]  \label{3.4.1}%
\end{equation}

\bigskip Using the D-H parameter table, homogeneous transformations matrixes
are resulted%
\begin{align}
H_{0}^{1} &  =%
\begin{bmatrix}
\cos\theta_{1} & -\sin\theta_{1} & 0 & 0\\
\sin\theta_{1} & \cos\theta_{1} & 0 & 0\\
0 & 1 & 0 & d_{1}\\
0 & 0 & 0 & 1
\end{bmatrix}
,\text{ \ \ \ \ }H_{1}^{2}=%
\begin{bmatrix}
\cos\theta_{2} & -\sin\theta_{2} & 0 & 0\\
0 & 0 & -1 & 0\\
\sin\theta_{2} & \cos\theta_{2} & 0 & 0\\
0 & 0 & 0 & 1
\end{bmatrix}
\text{\ \ , }\label{3.4.2}\\
H_{2}^{3} &  =%
\begin{bmatrix}
\cos\theta_{3} & -\sin\theta_{3} & 0 & \alpha_{3}\\
\sin\theta_{3} & \cos\theta_{3} & 0 & 0\\
0 & 1 & 0 & 0\\
0 & 0 & 0 & 1
\end{bmatrix}
\text{\ }\nonumber\\
H_{3}^{4} &  =%
\begin{bmatrix}
\cos\theta_{4} & -\sin\theta_{4} & 0 & \alpha_{4}\\
\sin\theta_{1} & \cos\theta_{1} & 0 & 0\\
0 & 1 & 0 & 0\\
0 & 0 & 0 & 1
\end{bmatrix}
,\text{ \ \ \ \ }H_{4}^{5}=%
\begin{bmatrix}
\cos\theta_{5} & -\sin\theta_{5} & 0 & 0\\
0 & 0 & 1 & 0\\
\sin\theta_{5} & \cos\theta_{5} & 0 & 0\\
0 & 0 & 0 & 1
\end{bmatrix}
\nonumber
\end{align}

\bigskip The multiplication of the matrixes (\ref{3.4.1})-(\ref{3.4.2}) gives
the table of the total homogeneous transformation, which is expressed:%
\begin{equation}
H_{0}^{5}=H_{0}^{1}\ast H_{1}^{2}\ast H_{2}^{3}\ast H_{3}^{4}\ast H_{4}^{5}%
\begin{bmatrix}
n_{x} & o_{x} & \alpha_{x} & d_{x}\\
n_{y} & o_{y} & \alpha_{y} & d_{y}\\
n_{z} & o_{z} & \alpha_{z} & d_{z}\\
0 & 0 & 0 & 1
\end{bmatrix}
\label{3.4.6}%
\end{equation}

where:

\begin{itemize}
\item $n=\left(  n_{x},n_{y},n_{z}\right)  ^{T}$ is the vector representing
the direction of the axis $\left(  O_{1},x_{1}\right)  $ in the coordinate
system $\left(  O_{0},x_{0},y_{0},z_{0}\right)  $

\item $o=\left(  o_{x},o_{y},o_{z}\right)  ^{T}$ is the vector representing
the direction of the axis $\left(  O_{1},y_{1}\right)  $ in the coordinate
system $\left(  O_{0},x_{0},y_{0},z_{0}\right)  $

\item $\alpha=\left(  \alpha_{x},\alpha_{y},\alpha_{z}\right)  ^{T}$ is the
vector representing the direction of the axis $\left(  O_{1},z_{1}\right)  $
in the coordinate system $\left(  O_{0},x_{0},y_{0},z_{0}\right)  $

\item $d=\left(  d_{x},d_{y},d_{z}\right)  ^{T}$ is the vector representing
the direction of the axis $\left(  O_{1},x_{1}\right)  $ is the vector
representing the joint's position
\end{itemize}

where:

$n_{x}$= $((c_{1}c_{2}c_{3}-c_{1}s_{2}s_{3})c_{4}+(-c_{1}c_{2}s_{3}-c_{1}%
s_{2}c_{3})s_{4})c_{5}+s_{1}s_{5}$

$n_{y}$ = $((s_{1}c_{2}c_{3}-s_{1}s_{2}s_{3})c_{4}+(-s_{1}c_{2}s_{3}%
-s_{1}s_{2}c_{3})s_{4})c_{5}-c_{1}s_{5}$

$n_{z}$= $((s_{2}c_{3}+c_{2}s_{3})c_{4}+(-s_{2}s_{3}+c_{2}c_{3})s_{4})c_{5}$

$o_{x}$ = $-((c_{1}c_{2}c_{3}-c_{1}s_{2}s_{3})c_{4}+(-c_{1}c_{2}s_{3}%
-c_{1}s_{2}c_{3})s_{4})s_{5}-s_{1}c_{5}$

$o_{y}$ = $-((s_{1}c_{2}c_{3}-s_{1}s_{2}s_{3})c_{4}+(-s_{1}c_{2}s_{3}%
-s_{1}s_{2}c_{3})s_{4})s_{5}-c_{1}c_{5}$

$o_{z}$= $(c_{2}c_{3}-s_{2}s_{3})s_{4}-((-s_{2}c_{3}-c_{2}s_{3})c_{4}$

$\alpha_{x}$ = $-(c_{1}c_{2}c_{3}-c_{1}s_{2}s_{3})s_{4}+(-c_{1}c_{2}%
s_{3}-c_{1}s_{2}c_{3})c_{4}$

$\alpha_{y}$ = $-(s_{1}c_{2}c_{3}-s_{1}s_{2}s_{3})s_{4}+(-s_{1}c_{2}%
s_{3}-s_{1}s_{2}c_{3})c_{4}$

$\alpha_{z}$ = $(c_{2}c_{3}-s_{2}s_{3})c_{4}-(s_{2}c_{3}+c_{2}s_{3})s_{4}$

$d_{x}$ = $(-(c_{1}c_{2}c_{3}-c_{1}s_{2}s_{3})s_{4}+(-c_{1}c_{2}s_{3}%
-c_{1}s_{2}c_{3})c_{4})d_{5}+(c_{1}c_{2}c_{3}-c_{1}s_{2}s_{3})a_{4}+c_{1}%
c_{2}a_{3}$

$d_{y}$ = $(-(s_{1}c_{2}c_{3}-s_{1}s_{2}s_{3})s_{4}+(-s_{1}c_{2}s_{3}%
-s_{1}s_{2}c_{3})c_{4})d_{5}+(s_{1}c_{2}c_{3}-s_{1}s_{2}s_{3})a_{4}+s_{1}%
c_{2}a_{3}$

$d_{z}$ = $(-(s_{2}c_{3}+c_{2}s_{3})s_{4}+(-s_{2}s_{3}+c_{2}c_{3})c_{4}%
)d_{5}+(s_{2}c_{3}-c_{2}s_{3})a_{4}+s_{2}a_{3}+d_{1}$ [6]

and \ $c_{i}=\cos\theta_{i}$ \ , $s_{i}=\sin\theta_{i}$ .

\bigskip

\subsection{ Inverse Kinematic Analysis}

\begin{figure}[!h]
	\centering
	\includegraphics[scale=0.5]{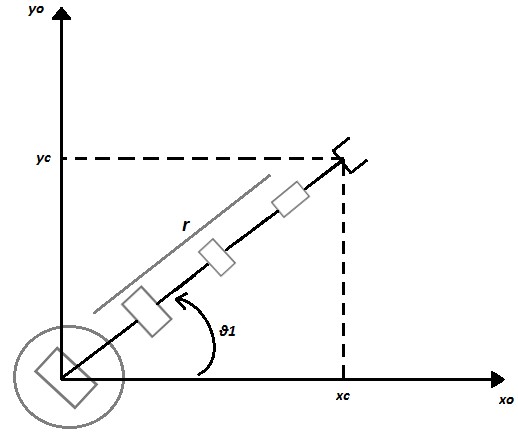}
	\caption{}
	\label{image315}
\end{figure}

In the Inverse kinematic (\cite{spong}, \cite{Graig2017})analysis we use the kinematics equations to find the
desired position of the end-effector. In other words, forward kinematics uses
the joint parameters to compute the configuration of a kinematic chain, and
inverse kinematics reverses this calculation to determine the joint parameters
that achieves a desired configuration. Calculation of the inverse kinematics
problem is much more complex than forward kinematics, since there is no unique
solution. In this project a geometric approach been used for solving the
inverse kinematics problem. The complexity of the inverse kinematic problem
increases with the number of nonzero link parameters and the geometric
approach that used to solve the problem is simplest and more natural. The
general idea of the geometric approach is to project the manipulator onto the
$x_{0}$-$y_{0}$ plane (Figure \ref{image315}) and solve a simple trigonometry problem to
find $\theta_{1}$

\bigskip

\bigskip

From the projection we can see that%
\begin{equation}
\theta_{1}=\text{atan}\left(  \frac{y}{x}\right)  \label{3.5.1}%
\end{equation}

The distance from the base to the edge of the grip is $r$, therefore%
\begin{equation}
r=\sqrt{x_{c}^{2}+y_{c}^{2}} \label{3.5.2}%
\end{equation}

where%
\begin{equation}
x_{c}=r\ast\cos(\theta_{1})\text{ \ and }y_{c}=r\ast\sin(\theta_{1})
\label{3.5.3}%
\end{equation}

\begin{figure}[!h]
	\centering
	\includegraphics[scale=0.5]{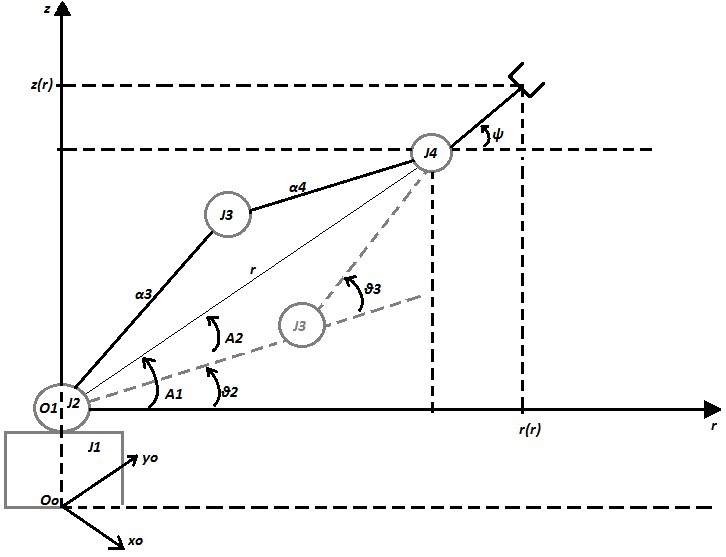}
	\caption{}
	\label{image352}
\end{figure}

A second projection of the manipulator is shown in Figure \ref{image352}. From this
projection we can see that the $A_{1}$ is :%
\begin{equation}
A_{1}=\text{atan}\left(  \frac{y}{x}\right)  \label{3.5.5}%
\end{equation}

The $A_{2}$ is :%
\begin{equation}
A_{2}=\frac{\text{acos(}a_{3}^{2}-a_{4}^{2}+r^{2})}{a_{3}^{2}\ast r}
\label{3.5.6}%
\end{equation}

Therefore, the angle of the shoulder joint $\theta_{2}$ is :%
\begin{equation}
\theta_{2}=A_{1}+A_{2} \label{3.5.7}%
\end{equation}

The second solution of the angle $\theta_{2}$ is written:%
\begin{equation}
\theta_{2}=A_{1}\pm A_{2} \label{3.5.8}%
\end{equation}

The elbow joint corresponds to the angle $\theta_{3}$ which equals to :%
\begin{equation}
\theta_{2}=\frac{\text{acos(}a_{3}^{2}+a_{4}^{2}+r^{2})}{a_{3}^{2}\ast a_{4}}
\label{3.5.9}%
\end{equation}

The relation between the angle $\psi$ (grip rotation) and the angles
$\theta_{1},\theta_{2},\theta_{3}$ is written:%
\begin{equation}
\theta_{4}=\psi-\theta_{2}-\theta_{3} \label{3.5.10}%
\end{equation}

\bigskip

From the geometric approach that analyzed above, the user will have complete
control of the manipulator, controlling all six arm servos. When the user
changes the angles $\theta_{2}$ and $\theta_{3}$, the angle $\theta_{4}$ will
not change, thus not changing the point of the grip and its orientation. This
is a consequence of the geometric approach (Figure \ref{image352}). In other words, the three thrust 
mechanisms $\theta_{4}$ (rotation of the wrist), $\theta_{5}$ (grip rotation),
$\theta_{6}$ (opening and closing of the grip) are not affected by the
movement of angular movement mechanisms $\theta_{1}$ (base rotation),
$\theta_{2}$(shoulder rotation) , $\theta_{3}$(elbow rotation), and the
inverse . [6]

\subsection{\bigskip SolidWorks CAD-CAE}

\begin{figure}[!h]
	\centering
	\includegraphics[scale=0.5]{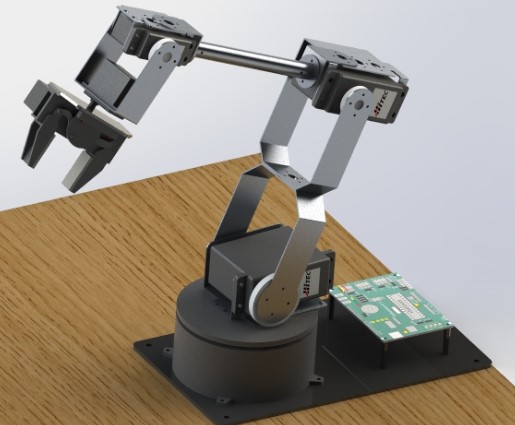}
	\caption{}
	\label{image361}
\end{figure}

All the parts comprising the project have been measured and designed with
usage of SolidWorks$^{\copyright}$ software. It's a solid modeling
computer-aided design (CAD) and computer-aided engineering (CAE) computer
program. Users can see the full dimensionality (2D or 3D) of every part
comprising the arm, as well as the material of every part. Additionally
SolidWorks provides users with \textquotedblleft countless\textquotedblright%
\ capabilities like measuring \ parts, mass properties, motion study,
collision check etc. In Figure \ref{image361}, we can see the final 3D rendering
of the arm.

\bigskip

\bigskip

\section{Electrical-Electronic wiring}

\begin{figure}[!h]
	\centering
	\includegraphics[scale=0.5]{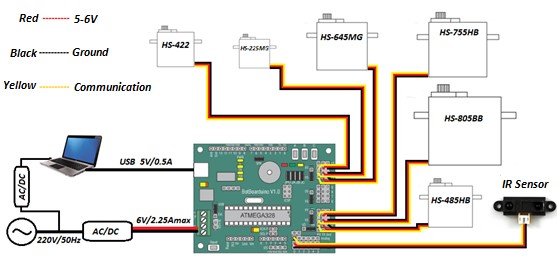}
	\caption{}
	\label{image41}
\end{figure}

Communication compatibility between devices and the proper powering of these
devices is important for the right operation of the project \cite{Forouzan}. Once the correct
devices have been selected based on their mechanical analysis, their
electrical behavior should be analyzed in order to avoid encounter any
communication or powering problems. In Figure \ref{image41} we can see the electrical
diagram of the project. A current source ($220VAC/50Hz$ socket) feeds the
computer's power supply adapter ($18.5VDC/6.5A$) and another source feeds the
BotBoarduino's power supply adapter \ ($6VDC/2.25A$). All servos are powered
by the \ BotBoarduino's adapter, the IR sensor is powered by the computer's
USB cable ($5VDC/0.5A$),through the BotBoarduino's regulator ($5VDC/1.5A$).
The red (+ positive) and black (-negative) cables of the servos and IR sensor
are for powering the devices, the yellow cable is for the communication with
microcontroller ATMEGA328 and the computer through the USB cable (see \cite{arduino},\cite{Automation},\cite{COMPANY},\cite{Chips},\cite{Lynxmotion},\cite{Hart}).

\bigskip

\bigskip

\subsection{Microcontroller Description}

\bigskip There are a lot of microcontroller boards in the market today with
different functions, depending on the needs. The microcontroller board that
was selected for this project is the BotBoarduino board because can give us
the desired currents for the devices (servos and IR sensor) and can also split
two sources of power for different powering needs. For example, in our
project, servos are powered with 6VDC by bridging the jumper to the VS input
and IR sensor is powered with 5VDC by bridging the jumper to the VL input.
BotBoarduino is based on the ATMEGA328 microcontroller, it has a USB mini port
that connects to a computer for programming the microcontroller. The
LD29150DT50R regulator that BotBoarduino has onboard can power up to 1.5A
current through the VL input. An external source of power (VS input) can also
be connected onto the board, like the adapter (6V/2.25A) we use, for powering
devices that need grater volts and amps, than the 5V/1.5A of VL input (see \cite{Automation},\cite{STMicroelectronics},\cite{Chips}). 

\bigskip

\subsection{Servomotors and IR Sensor}

\begin{figure}[!h]
	\centering
	\includegraphics[scale=0.5]{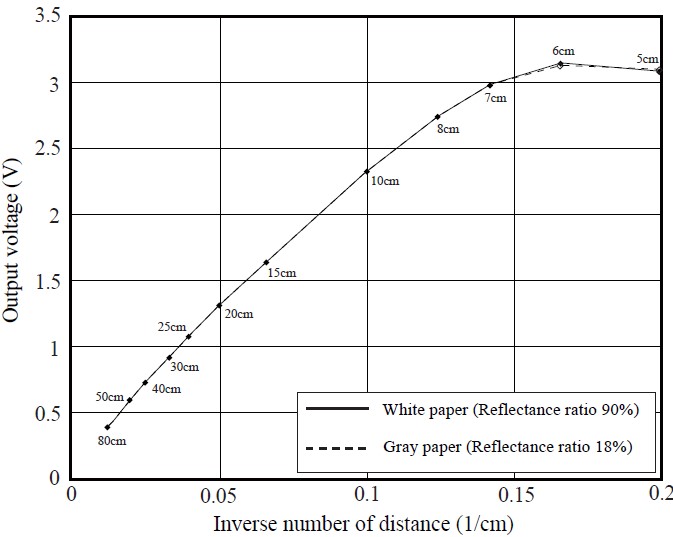}
	\caption{}
	\label{image421}
\end{figure}

The angle of each servo controlled by the Pulse Width Modulation (PWM) value
which is defined during programming. Each of the servo have an IC inside that
can check the emitted PWM of the BotBoarduino and drive the servo to the
desired position. Servos are powered by 6V DC by the adaptor through the
BotBoarduino. The distance of IR sensor is measured by the Position Sensitive
Detector (PSD) which then `'translate\textquotedblright\ the measured
distance, through the IC of the sensor, to an output voltage. The greater the
output voltage, the greater the distance. In the Figure \ref{image421} we can see the
curve of the measured distance in relation with the output voltage. We can
observe that the best accuracy region of the sensor is between the values of
10cm and 15cm. According to this, the distance between the sensor and the
sorting objects must be between these values, for best accuracy (see \cite{SHARP},\cite{Automation},\cite{COMPANY}).

\bigskip

\subsection{Power Consumptions}

\bigskip The power consumption of the devices of the arm, is one of the most
important topics in this study. The current consumed by the servos appears below:

\begin{itemize}
\item HS-485HB = 180mA

\item HS-805BB = 830mA

\item HS-755HB = 285mA

\item HS-645MG = 450mA

\item HS-225MG = 340mA

\item HS-422 = 180mA
\end{itemize}

The sum of the consumption current of the servos is 2.265mA. The adaptor we
selected for powering the servos gives us 2.250mA. The problem that occurs can
be solved by moving one servo at a time. We achieve that in the programming.
The communication current of each servo shown below:

\begin{itemize}
\item HS-485HB = 40mA

\item HS-805BB = 40mA

\item HS-755HB = 40mA

\item HS-645MG = 40mA

\item HS-225MG = 40mA

\item HS-422 = 40mA
\end{itemize}

The powering current of IR sensor is 30mA and the communication current is
40mA. The sum of the communication currents of the servos plus the powering
and communication currents of the sensor is 310mA. These currents powered by
the USB cable (5V-0.5A) of the computer through the `'LD29150DT50R current
regulator IC (5V-1.5A)\textquotedblright\ of the microcontroller. The
regulator gives us a maximum of 1.5A current, which means that we can overlap
the needs of 310mA (see \cite{arduino},\cite{Hart},\cite{STMicroelectronics}, \cite{Chips}).

\bigskip

\section{ Programming the Robotic Arm}

In this section we provide the programming of
the robotic arm. The software that have been used is the Arduino IDE (Integrated
Development Environment). The programming language that he software supports
are C and $C++$ (see \cite{Souli},\cite{Eckel}). Arduino IDE supplies programmers with software libraries
which provides them with many input or output procedures. A library, for
example, can be loaded into the program by writing \#include
%TCIMACRO{\TEXTsymbol{<}}%
%BeginExpansion
$<$%
%EndExpansion
math.h%
%TCIMACRO{\TEXTsymbol{>} }%
%BeginExpansion
$>$
%EndExpansion
to the command line of Arduino IDE, this library is for the identification of
the mathematical equations during programming. After the analysis of previous
studies (Mechanical/Electrical-Electronic Engineering) we came up with four
programs that show us the cooperation of the arm-sensor and its further
applications within contemporary industrial forms of production.

\bigskip

\subsection{Control cases}

\subsubsection{Autonomous Operation No.1}

\begin{figure}[!h]
	\centering
	\includegraphics[scale=0.5]{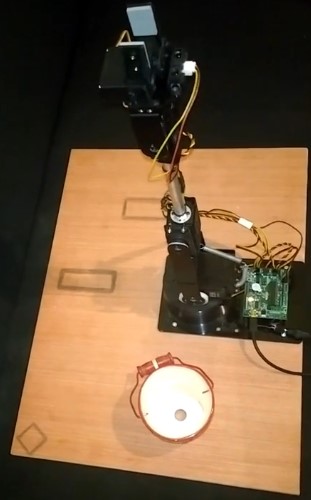}
	\caption{}
	\label{image511}
\end{figure}

\begin{figure}[!b]
	\centering
	\includegraphics[scale=0.65]{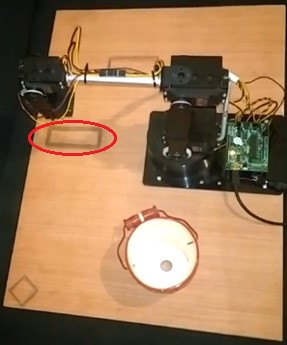}
	\caption{}
	\label{image512}
\end{figure}

During the first autonomous operation, the arm takes the initial start
position of the program, shown in Figure \ref{image511}, then the arm takes the
measuring position, according to the study that we have performed for the IR
sensor (Fig. \ref{image512}). As the arm reaches the measuring position, the IR sensor
starts to collect distance measurements between the `'eye\textquotedblright%
\ of the sensor and the sorting area, as shown in the red circle in Figure
\ref{image512}. The distance between the sensor and the empty sorting area is set to
13.8 cm approximately. If the measurement is lower than 13.8 cm the sensor is
set to recognize that an object is placed to the sorting area and the arm
picks it up and puts it into the bucket


\bigskip The video for the first autonomous operation can be seen in this URL: www.youtube.com/watch?v=srE3x6y4jqU\&feature=youtu.be

\bigskip

\subsubsection{Autonomous Operation No.2}

The initial start position and the measurement position is the same as the
autonomous operation No.1. If the sensor measures between
%TCIMACRO{\TEXTsymbol{<}}%
%BeginExpansion
$<$%
%EndExpansion
13.8cm and
%TCIMACRO{\TEXTsymbol{>}}%
%BeginExpansion
$>$%
%EndExpansion
=10cm, recognizes that the object is `'short\textquotedblright\ (predefined
measurements given). If the measurement is
%TCIMACRO{\TEXTsymbol{>}}%
%BeginExpansion
$>$%
%EndExpansion
greater than 10cm then the sensor recognizes that the object is `'tall\textquotedblright.
The arm, then picks the object and places it to a predefined positioned bucket
(Left for `'short\textquotedblright\ and Right for `'tall\textquotedblright).

The video for the second autonomous operation can be seen in this URL: www.youtube.com/watch?v=e8vaBb9g2A\&feature=youtu.be

\subsubsection{Autonomous Operation No.3}

\begin{figure}[!h]
	\centering
	\includegraphics[scale=1]{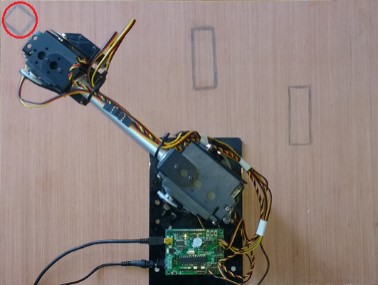}
	\caption{}
	\label{image5131}
\end{figure}

\begin{figure}[!h]
	\centering
	\includegraphics[scale=1]{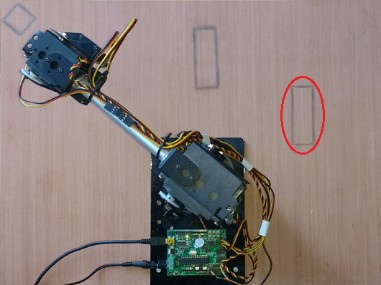}
	\caption{}
	\label{image5132}
\end{figure}

Third autonomous operation is almost the same \ as the autonomous operation
No.2. It differs in the placement of said objects. `'Short\textquotedblright%
\ object is placed to the predefined area, shown in Figure \ref{image5131} and
`'tall\textquotedblright\ object is placed to the predefined area, shown in
Figure \ref{image5132}


The video for the third autonomous operation can be seen in this URL: www.youtube.com/watch?v=rFqdlcLnQ08

\bigskip

\subsubsection{Manual Operation}

The fourth program is the manual operation of the arm performed by a user with
the keyboard of a computer loaded with said program. Users have full control
of the arm manipulating each servo independently. Additionally users can see
measurements of the sensor at the display of Arduino IDE.

The video for the manual operation can be seen in this URL:

www.youtube.com/watch?v=oFLjFvMqPjs\&feature=youtu.be ([18],[19],[20])

\bigskip

\section{Conclusions}

A robot arm has been designed to cooperate with an infrared sensor for the
identification of different sized objects and sorting them to predefined
positions. Also the manual operation of the arm through computer's keyboard is
presented. The study is based on the three main axes that Mechatronics consist:

\begin{itemize}
\item Mechanical Engineering

\item Electrical-Electronic Engineering

\item Computer Science
\end{itemize}

In the case of mechanical engineering, we analyze the torque of servos,
degrees of freedom and work space of the arm, mathematical modeling of forward
and inverse kinematics and the CAD of the arm. Analysis of
electrical-electronic engineering was important for the required powering of
the devices (BotBoarduino, Servos, IR Sensor), for communication between
devices and for the maximum efficiency of devices.
Three promising experiments have been conducted concerning the use of autonomous operations and the manual operation of the arm , that can be applied to the industry, as well as to other working environments. The IR sensor can identify a variety of objects based on their height and to validate position and orientation information of the grasped object. The designed robotic arm might be an educational one, but the procedure and the methodology followed is similar for an industrial type of robotic arm.
The next steps comprises, kinematic update to the arm, object reorientation routine,  Dynamics and kinematics of the object to improve stable grasping.

\newpage
%\bibliographystyle{99}
%\nocite{*}
%\bibliography{Robotics_biblio}

\end{document}